\documentclass{article}

\usepackage{arxiv}

\usepackage[utf8]{inputenc} 
\usepackage[T1]{fontenc}    
\usepackage{hyperref}       
\usepackage{url}            
\usepackage{booktabs}       
\usepackage{amsfonts}       
\usepackage{nicefrac}       
\usepackage{microtype}      
\usepackage{lipsum}
\usepackage{graphicx}
\usepackage{xcolor}
\usepackage{sistyle}
\usepackage{ifthen}
\usepackage[numbers]{natbib}
\SIthousandsep{,}

\title{From Same Photo: Cheating on Visual Kinship Challenges} 

\author{Mitchell Dawson \\
Visual Geometry Group (VGG) \\
Dept. of Engineering Science \\
University of Oxford \\
\texttt{mdawson@robots.ox.ac.uk} \\
\And
Andrew Zisserman \\
Visual Geometry Group (VGG)\\
Dept. of Engineering Science\\
University of Oxford \\
\texttt{az@robots.ox.ac.uk} \\
\And
Christoffer Nell\r{a}ker\\
Nuffield Dept. of Women's \\
\& Reproductive Health\\
Big Data Institute, IBME \\
University of Oxford \\
  \texttt{christoffer.nellaker@bdi.ox.ac.uk} \\
}

\begin{document}
\maketitle

\begin{abstract}
With the propensity for deep learning models to learn
unintended signals from data sets there is always the possibility that the network
can ``cheat'' in order to solve a task.
In the instance of data sets for visual kinship verification,  one such unintended signal could be that the faces are cropped from the same photograph, since faces from the same photograph are more likely to be from the same family. In this paper we investigate the influence of this artefactual data inference in published data sets for kinship
verification. 

To this end, we obtain a large data set, and train a CNN
classifier to determine if two faces are from the same photograph or not.
Using this classifier alone as a naive classifier of kinship, 
we demonstrate  near state of the art results  on five public benchmark data sets for kinship verification -- achieving over $90\%$ accuracy on one of them. Thus,  we conclude that faces derived from the same photograph  are  a strong inadvertent signal in all the data sets we examined, and it is likely that the fraction of kinship explained by existing kinship  models 
is small.
\end{abstract}

\section{Introduction}
\label{sec:intro}

Kinship verification is the task of determining whether two or more
people share a close family relation, using only photographs of their
respective faces. Potential uses of kinship verification include being
able to organise digital photo albums, detecting lost children using
photos of the child's parents, and in distinguishing family traits
from environmental or disease related changes in clinical research
contexts.

Over the last few years, several kinship verification data sets have
been developed for research groups to benchmark against. These 
data sets of images and videos have been produced by different
groups, and each has its own unique characteristics  including
variations in the  number of images, collection methods, and types of
relation included.

In building these data sets for kinship verification,  often different individuals' images have been cropped from larger family photos. While this is a good way to find people in the same family it also builds in another clue -- a bias that people from the same photo are related. This has been overlooked as a potential issue, as it is far easier to identify whether two face images  are from the same original photo than it is to determine if they are kin. As will be demonstrated, in some data sets, this `from same photograph' (FSP) signal can be used to achieve results comparable to the state of the art.

\begin{figure}[t]
\begin{center}
\includegraphics[width=0.9\textwidth]{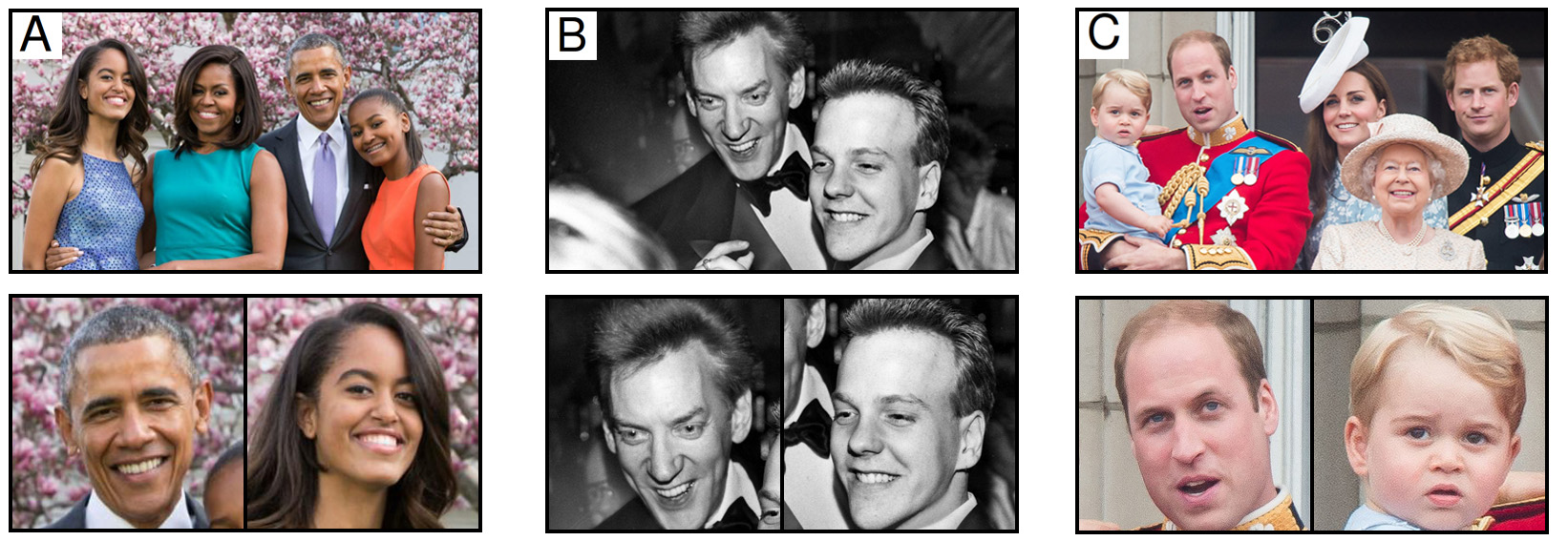}
\caption{Representative examples of how cropping face images from the same original photo can make the kinship verification a trivial task. Confounding, non--kinship information includes camera specific noise, background similarity (A,C), the historical era in which the photo was taken (B), image lighting and tone (A,B,C), and the relative age difference between parents and children (A,B,C).}
\vspace{-6mm}
\label{fig:FSP_family_examples}
\end{center}
\end{figure}

There are several cues that can be used to determine if two face
images are cropped from the same photograph, and by inference are more likely to be kin. For example, common lighting and shadows, background, blur, resolution, tone, contrast, clothing, camera specific noise and
exact regional overlap between crops. Another confounding signal,  which
is present when parent and child are cropped from the same photo, is
that the age difference between the people shown in each crop will be roughly the same as the average age which parents give birth to children.

Deep neural networks are notorious for finding subtle data shortcuts to exploit in
order to `cheat' and thus not learn to solve the task in the desired
manner; an example is the misuse of chromatic aberration
in~\citep{Doersch15} to solve the relative-position task. Thus
deep learning models are liable to to pick up on these FSP cues to gain a
boost in performance on kinship tasks. 

This problem was raised by Lopez {\it et al.}~\citep{lopez2016comments}, who
recommended that the two KinFaceW data sets should no longer be used in
kinship research. Their work showed that comparable results could be achieved 
on the data sets using a method with no understanding of
kinship: classifying pairs of images based upon the the distance
between the chrominance averages of images in the Lab color
space. Further to this work, Wu {\it et al.}~\citep{wu2016usefulness} showed
that results on kinship data sets can be improved by using colour
images rather than grey-scale images. Although one might expect that
skin colour could help to identify kinship relations, it also gives
more information about the background and colour distribution of the
image, a confounding variable which can be used to improve results on
data sets containing images cropped from the same photo.

Other groups have attempted to avoid this issue by using images  where
each family member is cropped from different original photographs,
such as the Family~$101$~\citep{fang2013kinship} and
UBKinFace~\citep{xia2011kinship} data sets. However, such images are
expensive to find and so photographs of famous people and their
children, who have many public images available, are frequently included,
introducing a new bias into the data sets.

Many models have already reported on these data sets infected by the signal of pairs cropped from the same photo. Therefore, we would like to benchmark the extent to which these results could have used the `from same photo' signal. This will provide a clear understanding of how much of the accuracy reported is based upon kinship signals, and how much could have been `from same photo' signal.

In this paper we benchmark the accuracy which can be achieved on five
popular kinship data sets, using only the signal of whether two face images
are from the same photo. We achieve this by creating a new data set of
24,827 images, containing 145,618 faces, and 914,517 pairs of faces
taken from non-familial group photos. We use this data set to train a
CNN classifier to determine if two facial images are
cropped from the same original photo or not. Crucially, this classifier has
no understanding of kinship relations.

We present results on the KinFaceW-I~\citep{lu2014neighborhood},
KinFaceW-II~\citep{lu2014neighborhood}, Cornell
KinFace~\citep{fang2010towards}, Families in the Wild
(FIW)~\citep{robinson2016families}, and TSKinFace~\citep{qin2015tri}
data sets. We show that the `from same photo' signal exists for all of these five data sets. For some, the signal is fairly weak, whereas for
others it explains a large proportion of the state of the art
results. In many cases we achieve comparable results with other
submissions using only the `from same photo' signal.

\vspace{-4mm}
\section{Training a Classifier to Detect Faces from the Same Photo}
\vspace{-4mm}

In this section, we describe our approach to train a CNN
classifier to be able to determine  whether two facial images are cropped
from the same original photograph or not.  For  positive training data we
use pairs of face images cropped from non-family group
photographs. This was done to ensure the classifier could not learn
kinship signals between faces. We evaluate the ability of our
classifier to classify faces from the same photo on a held out test
set.

\vspace{-5mm}
\subsection{Generating the FSP Data Set}\label{subsec:generate_FSP_data}

We would like to build a data set of face images that were cropped from the same original photo, but crucially where the people shown in the face images do not share a kinship relation. 

\begin{figure}[t]
\begin{center}
\includegraphics[width=0.9\textwidth]{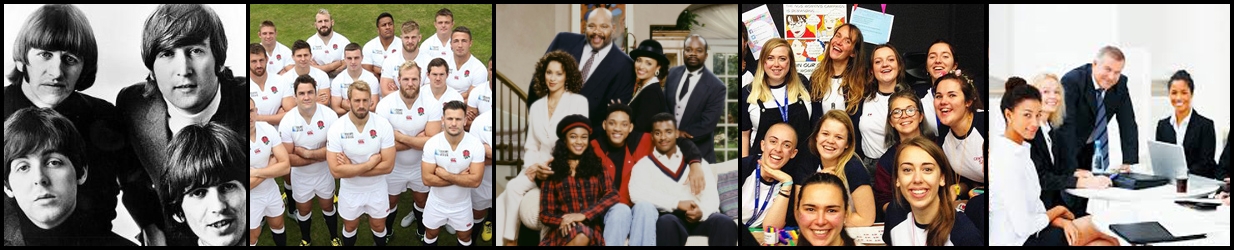}
\caption{Examples of photos found using non-kin group search terms and the Bing Azure API~\citep{BingImageAPI}. Pairs of faces found in these images are used in creating the From Same Photo (FSP) data set. }
\label{fig:FSP_group_examples}
\end{center}
\end{figure}

We began by creating a list of $125$ search terms that describe groups
of people, who are unlikely to share a kinship relation. These include
terms such as `school students', `business meeting', `sport team
photos' (the complete list of terms is given in the supplementary
material). We then use the Bing Azure API~\citep{BingImageAPI} to find
URLs of images related to these search terms. We ensure that the
images returned are large enough to potentially contain high
resolution faces, and that each URL is returned only once in an attempt
to avoid duplicate images. These searches result in a list of
$\num{81784}$ URLs: an average of 654 per search term. We
download each image and, with some loss due to moved content and unusable file formats, obtain $\num{76450}$ images.

\begin{figure}[!b]
\begin{center}
\includegraphics[width=0.9\textwidth]{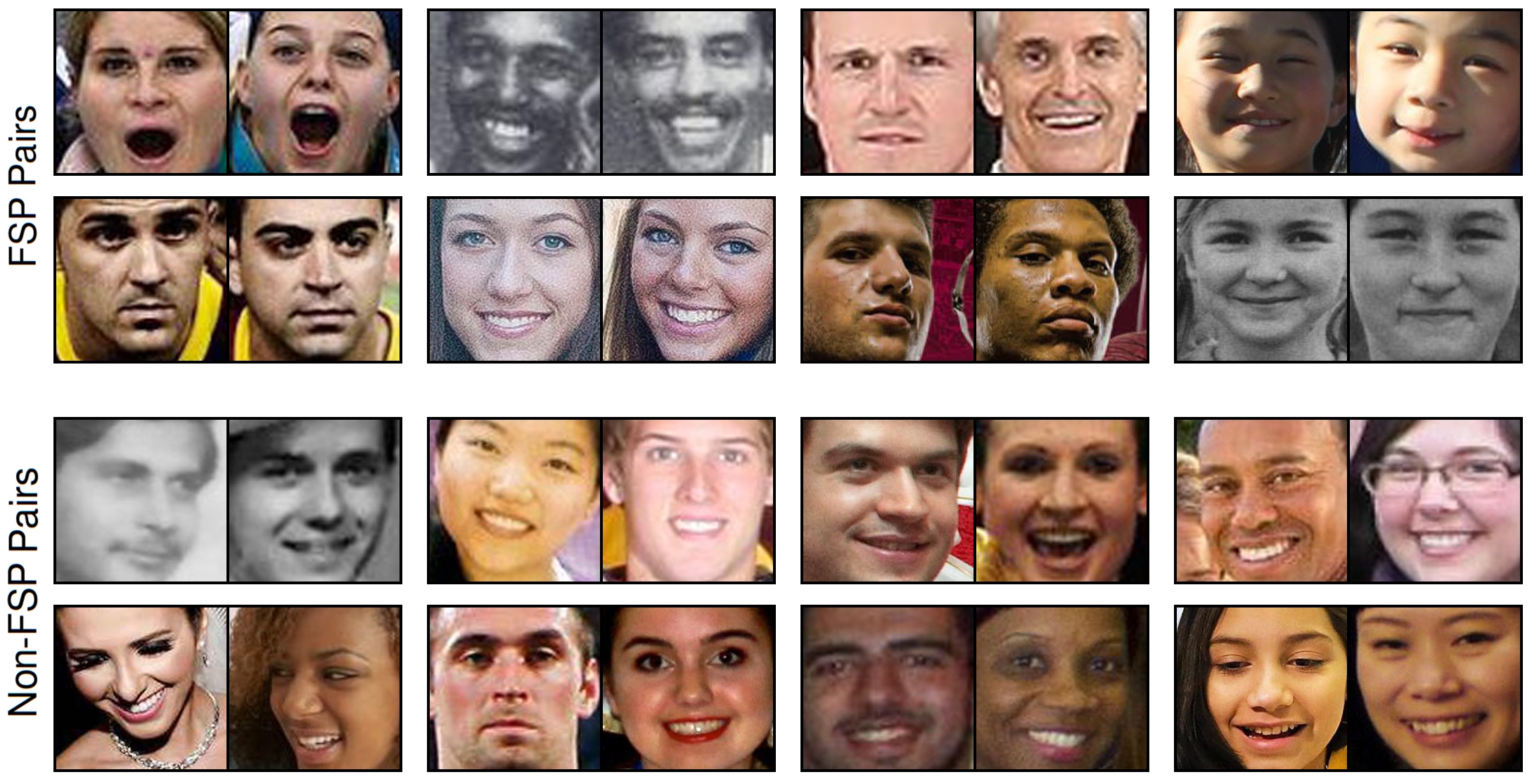}
\caption{Examples of pairs of faces used to train the From Same Photo (FSP) classifier. FSP pairs are cropped from the same non-kin group photo. Images in Non-FSP pairs are cropped from separate non-kin group photos}
\label{fig:FSP_train}
\end{center}
\end{figure}

The Dlib face
detector \citep{dlib09} is used to 
find  faces within these images. For our purposes, a usable group photo is
defined to be an image which contains at least two faces, where the
height and width of both face detection regions is not less than $50$
pixels. This results  in a maximum usable positive data set of
$\num{24827}$ group photos, which contains  $\num{145618}$ face images
and $\num{914517}$ possible pairs of face images. We partition this data set into training ($70\%$), validation ($10\%$) and testing ($20\%$),
as shown in Table~\ref{tbl:FSPdatasetANDresults}.

Although the vast majority of pairs of faces in the positive FSP data
set are correctly labelled, it was found that there are some cases of
negative pairs making it through the processing pipeline. For example,
some of the group images collected were composite collages of separate
photos, and so the faces in the collages were not originally taken
from the same photo. Another example which could lead to falsely
labelled positive pairs comes from photographs of people standing next
to portrait paintings or photographs which contain a face. These faces
are also detected by Dlib, and so lead to pairs of faces being falsely
labelled as positives. Furthermore, we can not exclude that some
images will have true kin related people present in them. Overall,
these examples make up a very tiny proportion of the data set, and so
it is not expected that they skew our results significantly.

Training the FSP classifier also requires creating a corresponding negative set of pairs of faces that are cropped from different original photographs. This is achieved by taking each image in the positive data set and randomly swapping one of the faces for another in the same training/validation/testing subset. This ensures that pairs of faces in the negative data set are not from the same photo, and that face photos do not leak across train, validation and testing splits. Furthermore, matching the total number of positive pairs and negative pairs in this way also helps to ensure that the FSP classifier does not learn a bias towards predicting a particular class.

\subsection{The FSP classifier}

\begin{figure}[b!]
\begin{center}
\includegraphics[width=0.9\textwidth]{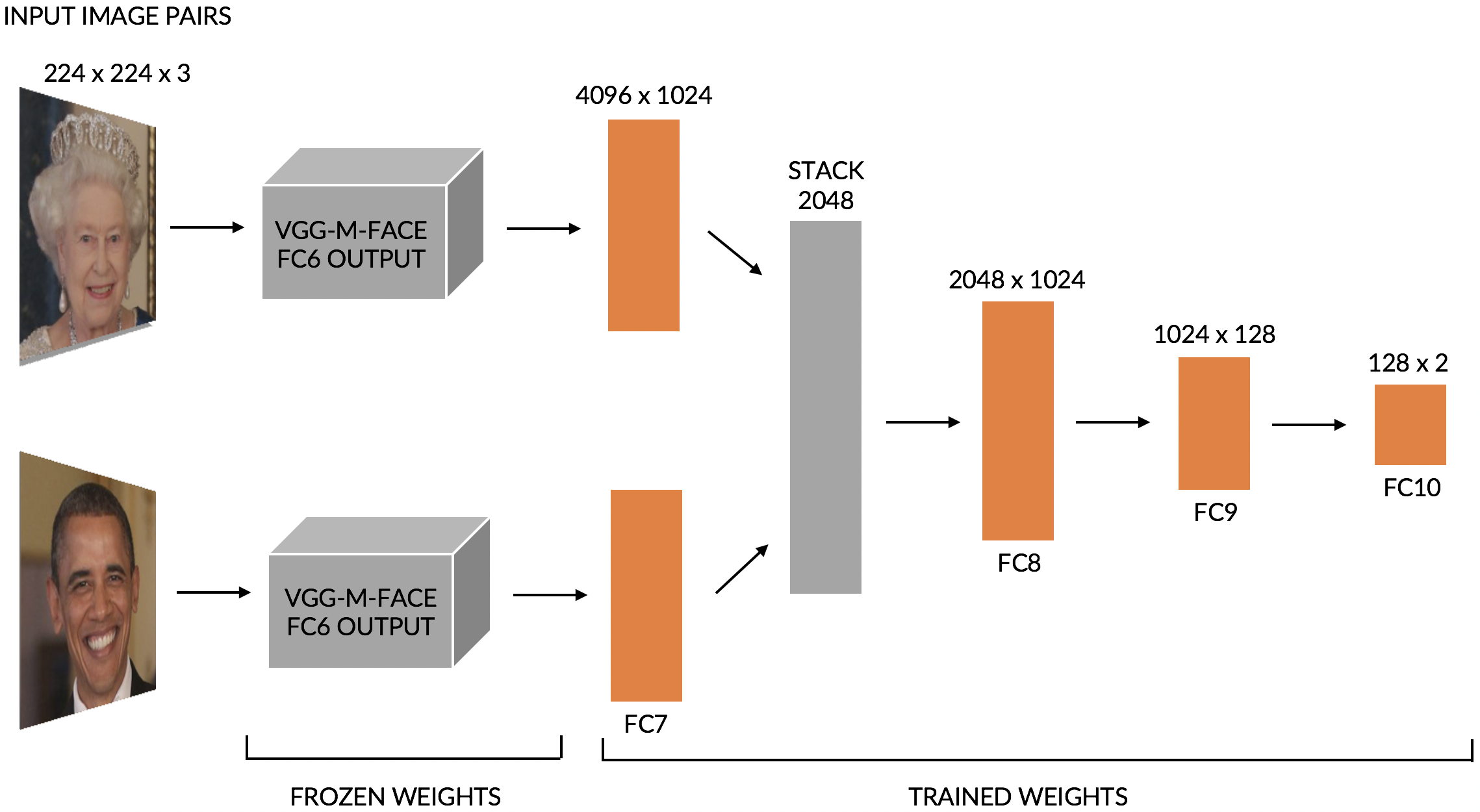}
\caption{The FSP network architecture.}
\label{fig:FSP_net_architechture}
\end{center}
\end{figure}

In this section we describe the CNN classifier used
to detect whether two facial photographs are cropped from the same
original image. For training data we use the balanced FSP data set
described above in section~\ref{subsec:generate_FSP_data}. We evaluate
the performance of the FSP classifier on a test set of images from the
FSP data set, and analyse the results using the standard receiver
operating characteristic (ROC) and precision-recall (PR) curves.

\paragraph{Architecture.} 
The CNN architecture can be split into four distinct parts as shown in
Figure~\ref{fig:FSP_net_architechture}. The first part consists of two
parallel copies of a pre-trained VGG-M-Face, up to the output from the
fully connected FC6 layer. This architecture was first used
in~\citep{chatfield2014return} and has been trained here for facial
recognition on the VGGFace data set. The 
VGG-M-Face networks generate
a $\num{4096}$-dimensional feature vector for each face image of the input pair. Both feature vectors are
then fed to a trainable fully connected layer to reduce their
dimensionality. These reduced feature vectors are then stacked into a
single vector and fed through through three more fully connected
layers and ReLU activation functions to produce a two-dimensional
vector giving the probability that the pair belongs to the class `from
same photo' or `not from same photo'. The network is implemented in
the PyTorch framework, and the code (and data set) will be made publicly
available. 

\begin{table}[t]
\begin{center}
\caption{Summary of the number of pairs of face images in the data sets used to train, validate and test the FSP classifier. The area under receiver operating characteristic curve (ROC AUC) is reported for the FSP classifier on the test set of images}
\begin{tabular}{ccccc}
\textbf{Data set} & \textbf{Train} & \textbf{Validation} & \textbf{Test}  & \textbf{ROC AUC (\%)} \\ \hline
\textit{FSP} & $\num{1321440}$ & $\num{184358}$ & $\num{323236}$ & $98.1$ \\ \hline
\end{tabular}
\end{center}
\label{tbl:FSPdatasetANDresults}
\end{table}

\begin{figure}[t]
\begin{center}
\includegraphics[width=0.9\textwidth]{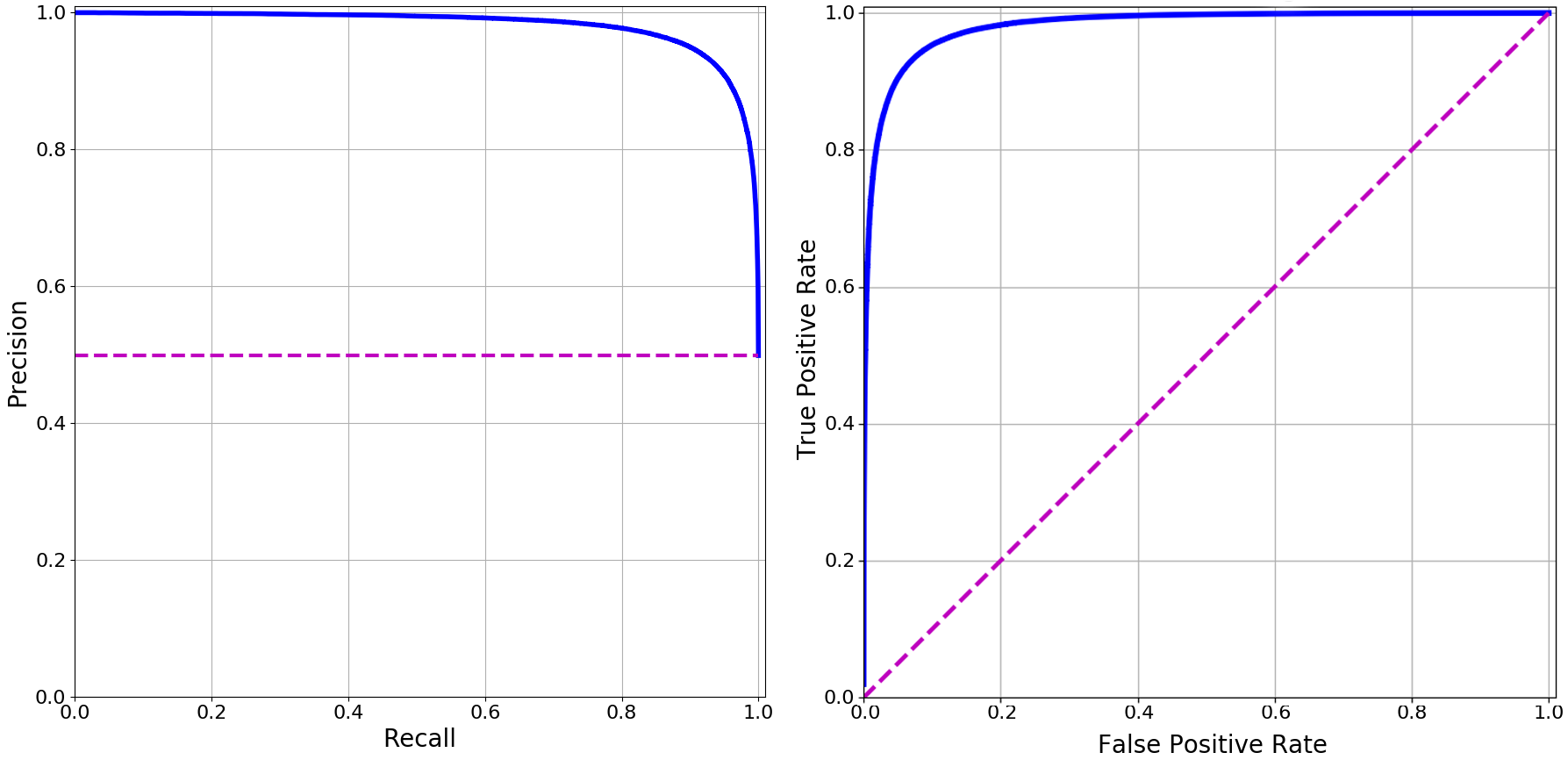}
\caption{Precision-recall and ROC curves  for testing the FSP classifier on the FSP test set. 
The area under the ROC curve is $98.09\%$, and the classifier performs with an equal error rate of $6.97\%$ and an average precision of $98.07\%$.}
\label{fig:PR_ROC}
\end{center}
\end{figure}

\paragraph{Training.} The FSP network was trained using stochastic gradient descent to minimise a softmax cross entropy loss function. During training we freeze the weights of the VGG-M-Face part of the model, but train all other layers. The initial learning rate was set to $0.1$, and decreased by a factor of 0.5 every five epochs, until the classifier's performance on the validation set began to saturate. In training we use dropout with $p=0.1$ between layers FC8 and FC9 to prevent overfitting. During training we augment our input images by resizing them up to $256\times256$ pixels and taking a random $224\times224$ pixel crop from the resulting image.

\paragraph{Testing.}  We evaluate the FSP network on a test set taken from the FSP data set that was not used in training or validation. The testing set consists of $\num{323236}$ pairs of face images, half of which are cropped from the same original photo. At test time we test a pair of images ten times using different permutations of region crops and horizontal flips. The result of these ten tests is then averaged to give the classifier score for a pair of images.

As can be seen in Figure~$5$, the FSP classifier is able
to tell whether two images are from the same photograph with very high
accuracy. Testing resulted in an area under ROC curve of $98.09\%$, with
an equal error rate of $6.97\%$.

\section{Testing FSP Classifier on Kinship Data Sets}

Kinship verification tasks have branched into three major challenges, consequently multiple different types of kinship verification data sets have been built. The first type of task is one-to-one kinship verification, where one wishes to determine whether two individuals are related. Alternatively, tri-subject kinship verification is where given a picture of a mother-father and a third person, we wish to determine if the third person is the biological child of the two parents. A final challenge is one-to-many family classification, where one wishes to know which family a person belongs to among a data set. We examined five popular data sets and the degree to which FSP is a biasing factor for the respective one-to-one kinship verification task challenges.

\subsection{Kinship Verification Data Sets}

\paragraph{KinFaceW-I \& KinFaceW-II (Lu et al.~\citep{lu2014neighborhood}).} 
These are two widely used kinship verification data sets 
introduced in 2014. They contain images of
public figures paired with their parents or children. They are the most
commonly tested on kinship data sets, but were also the first to receive criticism for using images cropped from the same photograph~\citep{lopez2016comments}. The KinFaceW-I data set contains pairs of
images from the four main bi-subject kinship verification categories:
mother-daughter (127 pairs), mother-son (116 pairs), father-daughter
(134 pairs), father-son (156 pairs). The major differences between the
two data sets are that KinFaceW-II is larger, and contains a greater
number of photos cropped from the same original image compared to
KinFaceW-I.

Typically these data sets are tested using five-fold cross validation,
where both the positive and negative pairs are specified for each of
the five folds. Here we collect all the positive and negative pairs
across all five folds to create the balanced test set. The KinFaceW-I
test set contains $\num{1066}$ pairs of face images, and the
KinFaceW-II test set contains $\num{2000}$ pairs of face images, with
250 for each of the four major relations.

\paragraph{Cornell KinFace (Fang et al.~\citep{fang2010towards}).}
This data set was published in 2010 and consists of parents
and children of public figures and celebrities. It is the smallest
data set among the five tested in this paper, consisting of only $144$
positive pairs. Human level performance on this data set was
benchmarked as $70.7\%$. For the negative set we randomly substitute 
a parent or child from the positive pair with a random parent or child
from the remaining positive set. To avoid bias from choosing a
particularly  easy negative set we average over five different versions
of the negative set, randomised uniquely each time.

\paragraph{Families in the Wild (FIW) (Robinson et al.~\citep{robinson2016families}).} 
Families in the wild is by far the largest of the data sets we test
on. It contains $\num{1000}$ families, $\num{10676}$ people and
$\num{30725}$ face images. $\num{418060}$ pairs of images, heavily
weighted  towards  over $\num{37000}$ pairs for each of the four major
parent-child relationships, and over $\num{75000}$ for sibling-sibling
relationships. The data set was made available as part of the 2018
Recognising Families in the Wild (RFIW) kinship verification
challenge. Here we test on the $\num{99962}$ pairs of images in the challenge evaluation set.

\paragraph{TSKinFace (Qin et al.~\citep{qin2015tri}).}
TSKinFace was introduced
as an alternative to the growing number of bi-subject
kinship verification data sets. For many of the use
situations described for kinship verification, such as organising
family photo albums or locating missing children, it is likely that
pictures of both parents will be available. TSKinFace is the largest
publicly available data set of triplet kinship images to date. The
data set contains 787 images of $\num{2589}$ individuals belonging to $\num{1015}$
tri-subject kinship groups (father-mother-child). The images from TSKinFace
are collected from family photographs.

\begin{table}[b!]
\begin{center}
\caption{Results of using the FSP classifier to predict kinship on KinFaceW-I, KinFaceW-II, Cornell KinFace, FIW and TSKinFace data sets. Accuracy percentages are shown for the pair subsets of mother-daughter (MD), mother-son (MS), father-daughter (FD), and father-son (FS), and the triplet subsets of father-mother-daughter (FMD) and father-mother-son (FMS), as well as across the entire test set.}
\label{tbl:FSP_results_breakdown}
\begin{tabular}{|l|c|c|c|c|c|c|c|}
\hline
Data set & MD & MS & FD & FS & FMD & FMS &  All\\
\hline\hline
KinFaceW-I & 86.0 & 78.3 & 74.1 & 74.6 & & & 76.8\\
KinFaceW-II & 94.8 & 90.3 & 84.5 & 92.3 & & & 90.2\\
FIW & 60.3 & 59.3 & 59.0 & 57.5 & & & 58.6 \\ 
TSKinFace  &  &  &  & & 88.6 & 89.4 & 88.6 \\
Cornell KinFace &  &  &  & & & & 76.7 \\
\hline
\end{tabular}
\end{center}
\end{table}

TSKinFace specifies related triplets of father-mother-son and
father-mother-daughter. For the negative set they randomly combine
images of parents with an image of different child of the same gender
as their own. To reduce the probability of reporting results on a
particularly easy negative set, we repeated our experiments with a
different set of permutations for the children and parents in the
negative set each time. The data set is split by the gender of the
child with 513 positive father-mother-son (FM-S) triplets and 502
positive father-mother-daughter (FM-D) triplets.

At test time we split each triplet into two pairs, father-child and
mother-child. Each pair is then scored by the FSP classifier. We take
the maximum of these two scores as the test score for
a triplet. This corresponds to asking whether at
least one of the parents is cropped from the same original photo as the
child.

\subsection{Implementation Details}

We apply the FSP classifier to the kinship data sets naively, that is to say without further training for the intended kinship verification task. To achieve this we extract each image at three crop sizes. This is required as the Dlib face detector tends to propose tight square regions around the centre of a face, often cutting out chin, tops of heads and ears. However, the face images contained within two of the data sets we test on are more loosely cropped. We found that the greatest accuracy was achieved on KinFaceW-I, KinFaceW-II and FIW using the FSP classifier trained with images cropped to the standard Dlib size. Whereas on Cornell KinFace and TSKinFace, the best results were obtained by expanding the width and height of
the images by $15\%$.

We report the accuracy our FSP classifier is able to obtain on each
kinship test set. To determine the accuracy, we set the threshold of our
FSP classifier using five-fold cross validation on the test sets. We
determine the threshold which produces the maximum achievable accuracy 
for the FSP classifier on $80\%$ of the test set, then calculate the accuracy 
on the remaining $20\%$ of the test data for five splits. The mean accuracy 
across the splits is then reported.

\section{Results} 

\begin{figure}[!b]
\begin{center}
\includegraphics[width=0.94\textwidth]{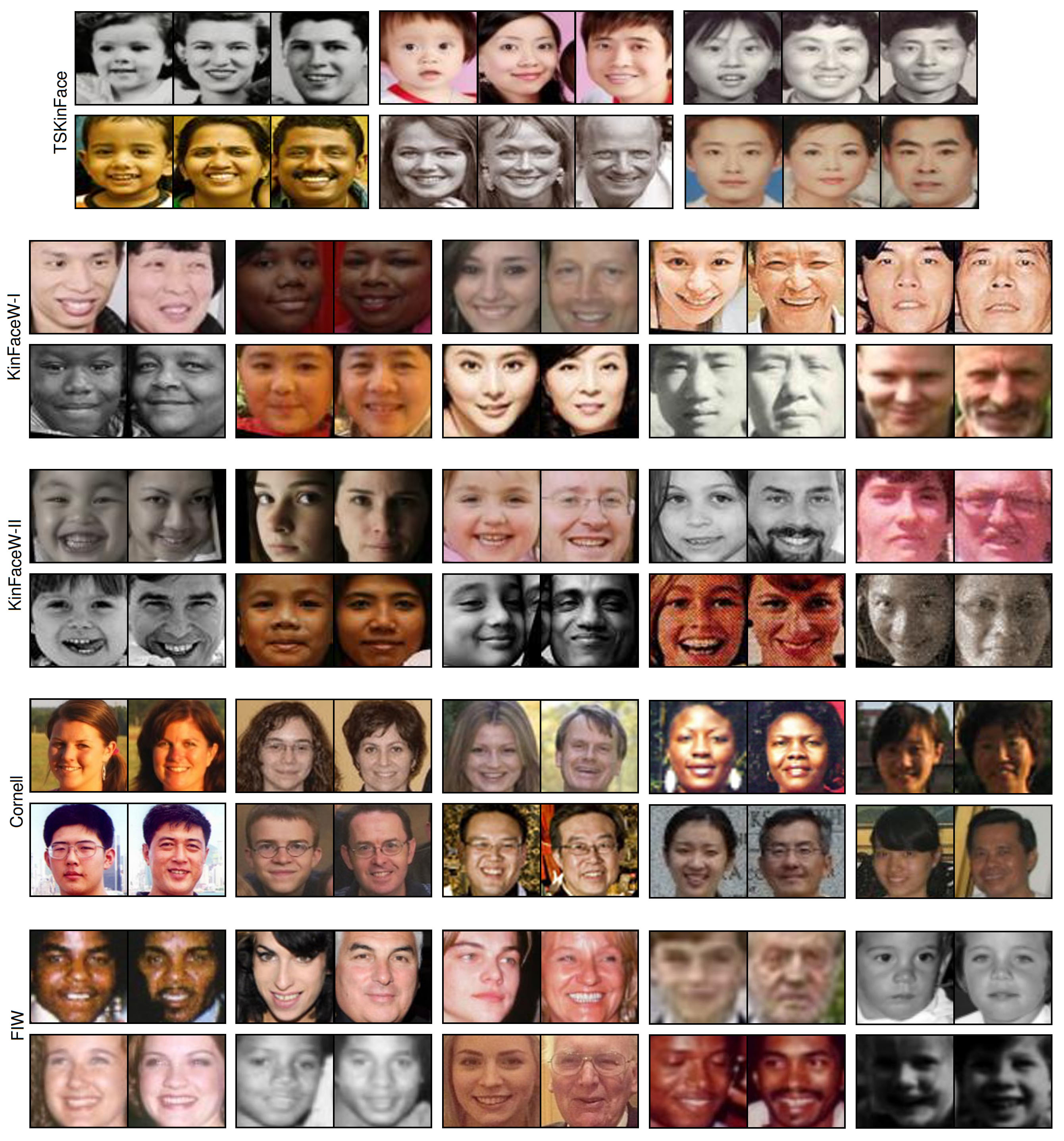}
\caption{Examples found in kinship data sets which we predict are taken from the same photo. Each row corresponds to examples from one of the five analysed kinship  data sets. Note the similarities in background, facial pose, lighting, and overall image tone between images cropped from the same original photo.}
\label{fig:FSP_family_examples_3}
\end{center}
\end{figure}

We show the results for the FSP classifier as a kinship classifier on
the five data sets in Table~\ref{tbl:FSP_results_breakdown}.
For each of the bi-subject kinship verification tasks we report high
accuracies: KinFaceW-I $76.8\%$ , KinFaceW-II 90.2\%, FIW 58.6\%,
Cornell KinFace 76.7\%. There is variance between the tasks depending
on the gender of the parent and child in the image. In kinship
verification tasks this could be interpreted as perhaps biases in
facial similarities between genders, however this is not something we
should expect from the FSP classifier. It is more likely to be due to
biases in likelihood for various gender pairs of family members to
show up together in photographs, or even just stochastic variance
from data set sampling.

\begin{table}[t]
\begin{center}
\caption{Results of using the FSP classifier to predict kinship, and previously published kinship classifiers
on all five kinship data sets. KinFaceW is abbreviated as KFW. Note the high
accuracies achieved by using the FSP classifier as a kinship
classifier on each data set, in comparison with the average and
state-of-the-art maximum accuracies over results reported in previous publications. Also note the available human annotation benchmarks
which are all exceeded by the FSP classifier model}
\label{tbl:results_comparison}
\begin{tabular}{|l|c|c|c|c|c|c|}
\hline
Paper & Year & KFW-I & KFW-II & Cornell KF & FIW & TSKinFace \\
\hline\hline
FSP classifier (ours) & 2018 & 76.8 & 90.2 & 76.7 & 58.6 & 88.6 \\
Mean Accuracy & & 78.3 & 82.0 & 79.2 & 67.2 & 85.7  \\
Median Accuracy & & 78.8 & 82.8 & 79.0 & 68.8 & 87.2  \\
Max Accuracy & & 96.9 & 97.1 & 94.4 & 74.9 & 93.4 \\
Human~\citep{zhao2018learning,fang2010towards,qin2015tri} & & 71.0 & 74.0 & 67.2 & & 79.5\\
\hline
Aliradi et al.~\citep{aliradi2018dieda} & 2018 & 80.6 & 88.6 & & & \\
Moujahid et al.~\citep{moujahid2018pyramid} & 2018 & 88.2 & 88.2 & & & \\
Lopez et al.~\citep{lopez2018kinship} & 2018 & 68.4 & 66.5 & & & \\
Yan et al.~\citep{yan2018learning} & 2018 & 77.6 & 78.5 & & & \\
Robinson et al.~\citep{robinson2018visual} & 2018 & 82.4 & 86.6 & & & \\
Kohli et al.~\citep{kohli2018supervised} & 2018 & 96.9 & 97.1 & 94.4 & & \\
Zhou et al.~\citep{zhou2018learning} & 2018 & 82.8 & 85.7 & 81.4 & & \\
Mahpod et al.~\citep{mahpod2017kinship} & 2018 & 79.8 & 87.0 & 76.6 & & \\
Zhao et al.~\citep{zhao2018learning} & 2018 & 81.5 & 82.5 & 81.7 & & 84.5 \\
Wang et al.~\citep{wang2018cross} & 2018 & & & & 69.5 & \\
Xia et al.~\citep{xia2018graph} & 2018 & & & & & 90.7 \\
Yang et al.~\citep{yang2017novel} & 2017 & 88.6 & 90.3 & & & 93.4 \\
Chen et al.~\citep{chen2017kinship} & 2017 & 83.3 & 84.3 & & & \\
Lu et al.~\citep{lu2017discriminative} & 2017 & 83.5 & 84.3 & & & \\
Patel et al.~\citep{patel2017evaluation} & 2017 & 78.7 & 80.6 & & & \\
Kohli et al.~\citep{kohli2017hierarchical} & 2017 & 96.1 & 96.2 & 89.5 & & \\
Laiadi et al.~\citep{laiadi2017rfiw} & 2017 & & & 83.2 & 54.8 & \\
Duan et al.~\citep{duan2017advnet} & 2017 & & & & 66.6 & \\
Dahan et al.~\citep{dahan2017kin} & 2017 & & & & 65.0 & \\
Li et al.~\citep{li2017kinnet} & 2017 & & & & 74.9 & \\
Wang et al.~\citep{wang2017kinship} & 2017 & & & & 68.8 & \\
Zhou et al.~\citep{zhou2016ensemble} & 2016 & 78.8 & 75.7 & & & \\
Xu et al.~\citep{xu2016kinship} & 2016 & 77.9 & 77.1 & & & \\
Liu et al.~\citep{liu2016novel} & 2016 & 77.9 & 81.4 & 75.9 & & \\
Zhang et al.~\citep{zhang2016genetics} & 2016 & & & &  & 89.7 \\
Robinson et al.~\citep{robinson2016families} & 2016 & & & & 71.0 & \\
Duan et al.~\citep{duan2015feature} & 2015 & 73.3 & 75.2 & & & \\
Bottinok et al.~\citep{bottinok2015multi} & 2015 & 86.3 & 83.1 & & & \\
Zhang et al.~\citep{zhang122015kinship} & 2015 & 77.5 & 88.4 & & & \\
Kou et al.~\citep{kou2015learning} & 2015 & 63.5 & 68.9 & & & \\
Yan et al.~\citep{yan2015prototype} & 2015 & 70.1 & 77.0 & 71.9 & & \\
Zhang et al.~\citep{zhang2015group} & 2015 & 67.0 & 86.0 & & & 89.0 \\
Qin et al.~\citep{qin2015tri} & 2015 & & & & & 85.4 \\ 
Lu et al~\citep{lu2014neighborhood} & 2014 & 69.9 & 76.5 & 66.5 & & 72.1 \\
Dehghan et al.~\citep{dehghan2014look} & 2014 & & & & & 80.8 \\
Nguyen et al. ~\citep{nguyen2010cosine} & 2011 & 62.5 & 71.9 & & & \\
Fang et al.~\citep{fang2010towards} & 2010 & & & 70.7 & & \\
Weinberger et al.~\citep{weinberger2009distance} & 2009 & 63.3 & 74.5 & & & \\
\hline
\end{tabular}
\end{center}
\end{table}

In Table~\ref{tbl:results_comparison}, we show summary statistics across a large number of published kinship verification models on the various data sets, and show that the FSP classifier performs kinship classification as well or better than most of them. We also achieve high accuracies for the tri-subject verification data
set beating many published models and near the state-of-the-art
published results (Table~\ref{tbl:results_comparison}, TSKinFace
88.6\%). In this instance there is no significant gender bias in
accuracy of verification of kinship for the gender of the child.
Finally, we
can see that the FSP classifier consistently outperforms human
classification of kinship. We can expect that the deep learning models 
will have  inadvertently learnt the FSP signal as a means to solve
the kinship verification task,  but can only speculate if human
classifiers uses the same type of information to simplify the task. Examples of positive kin pairs and triplets found by the FSP classifier are shown in Figure~$6$. A selection of pairs the classifier falsely predicted as being from the same photo are displayed in Figure~$7$.

\begin{figure}[t]
\begin{center}
\includegraphics[width=0.9\textwidth]{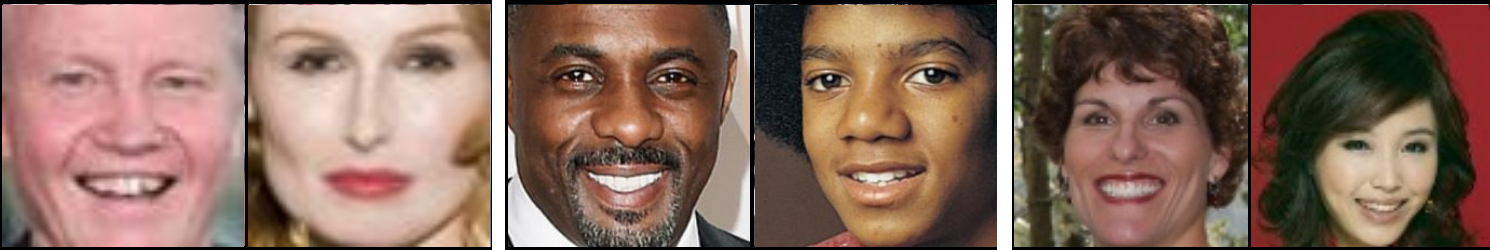}
\caption{Examples of non-kinship pairs that the classifier incorrectly predicted as being from the same photo. Pairs shown, from left top right, are from the KinFaceW-I, Families in the Wild and Cornell KinFace data sets respectively.}
\label{fig:FSP_family_examples_2}
\end{center}
\end{figure}

\section{Conclusions}

In this work, we have applied a `from same photograph' (FSP)
classifier as a naive classifier of kinship to five data sets regularly used for kinship verification tasks, and thereby have estimated the degree to which the FSP signal contaminates these data sets. The FSP classifier performs amongst the best published kinship classifier models, despite not being trained for this  primary task. It is likely that deep neural network models  built with the intention to verify kinship, are instead primarily using this much easier to detect FSP signal. We have also obtained a new data set for training classifiers to detect facial crops from the same photograph to begin to address this problem.

Furthermore, it is important to note that there are many other spurious signals that one should expect in  kinship verification data sets. For instance, deep learning kinship classifiers would also be expected to learn biases in distributions of age, gender and particularly ancestral backgrounds. Due to the way the FSP classifier has been trained it is blind to many of these other confounding non-kinship signals contained within existing kinship data sets. 

We recommend that an FSP classifier should be an important part of kinship data set production pipelines: either to ensure the FSP signal is removed entirely, or is balanced between positive and negative kin pairs. It should be considered that inappropriate construction of negative sets can introduce biases by overly simplifying the task, such as generating father-father-child tri-subjects, mismatched ancestral backgrounds, and implausible age distributions (such as a child older than the parents). Kinship verification is a challenging task made more difficult by inherent biases in how data occurs in the wild. Sharing a photograph does not make us relatives, but we are likely to share a photograph if we are.

\section*{Acknowledgments}
This research was financially supported by the EPSRC programme grant Seebibyte EP/M013774/1, the EPSRC Systems Biology DTC EP/G03706X/1, and the MRC Grant MR/M014568/1.

\bibliographystyle{plainnat}
\bibliography{references}

\begin{thebibliography}{46}
\providecommand{\natexlab}[1]{#1}
\providecommand{\url}[1]{\texttt{#1}}
\expandafter\ifx\csname urlstyle\endcsname\relax
  \providecommand{\doi}[1]{doi: #1}\else
  \providecommand{\doi}{doi: \begingroup \urlstyle{rm}\Url}\fi

\bibitem[Aliradi et~al.(2018)Aliradi, Belkhir, Ouamane, and
  Elmaghraby]{aliradi2018dieda}
Rachid Aliradi, Abdelkader Belkhir, Abdelmalik Ouamane, and Adel~S Elmaghraby.
\newblock Dieda: discriminative information based on exponential discriminant
  analysis combined with local features representation for face and kinship
  verification.
\newblock \emph{Multimedia Tools and Applications}, pages 1--18, 2018.

\bibitem[Bottinok et~al.(2015)Bottinok, Islam, and Vieira]{bottinok2015multi}
Andrea Bottinok, Ihtesham~Ul Islam, and Tiago~Figueiredo Vieira.
\newblock A multi-perspective holistic approach to kinship verification in the
  wild.
\newblock In \emph{Automatic Face and Gesture Recognition (FG), 2015 11th IEEE
  International Conference and Workshops on}, volume~2, pages 1--6. IEEE, 2015.

\bibitem[Chatfield et~al.(2014)Chatfield, Simonyan, Vedaldi, and
  Zisserman]{chatfield2014return}
Ken Chatfield, Karen Simonyan, Andrea Vedaldi, and Andrew Zisserman.
\newblock Return of the devil in the details: Delving deep into convolutional
  nets.
\newblock \emph{arXiv preprint arXiv:1405.3531}, 2014.

\bibitem[Chen et~al.(2017)Chen, An, Yang, and Wu]{chen2017kinship}
Xiaojing Chen, Le~An, Songfan Yang, and Weimin Wu.
\newblock Kinship verification in multi-linear coherent spaces.
\newblock \emph{Multimedia Tools and Applications}, 76\penalty0 (3):\penalty0
  4105--4122, 2017.

\bibitem[Corporation(2018)]{BingImageAPI}
Microsoft Corporation.
\newblock Bing {I}mage {S}earch {API}, 2018.
\newblock
  \url{https://azure.microsoft.com/en-us/services/cognitive-services/bing-image-search-api/}.

\bibitem[Dahan et~al.(2017)Dahan, Keller, and Mahpod]{dahan2017kin}
Eran Dahan, Yosi Keller, and Shahar Mahpod.
\newblock Kin-verification model on fiw dataset using multi-set learning and
  local features.
\newblock In \emph{Proceedings of the 2017 Workshop on Recognizing Families In
  the Wild}, pages 31--35. ACM, 2017.

\bibitem[Dehghan et~al.(2014)Dehghan, Ortiz, Villegas, and
  Shah]{dehghan2014look}
Afshin Dehghan, Enrique~G Ortiz, Ruben Villegas, and Mubarak Shah.
\newblock Who do i look like? determining parent-offspring resemblance via
  gated autoencoders.
\newblock In \emph{Proceedings of the IEEE Conference on Computer Vision and
  Pattern Recognition}, pages 1757--1764, 2014.

\bibitem[Doersch et~al.(2015)Doersch, Gupta, and Efros]{Doersch15}
Carl Doersch, Abhinav Gupta, and Alexei~A. Efros.
\newblock Unsupervised visual representation learning by context prediction.
\newblock In \emph{International Conference on Computer Vision (ICCV)}, 2015.

\bibitem[Duan and Zhang(2017)]{duan2017advnet}
Qingyan Duan and Lei Zhang.
\newblock Advnet: Adversarial contrastive residual net for 1 million kinship
  recognition.
\newblock In \emph{Proceedings of the 2017 Workshop on Recognizing Families In
  the Wild}, pages 21--29. ACM, 2017.

\bibitem[Duan and Tan(2015)]{duan2015feature}
Xiaodong Duan and Zheng-Hua Tan.
\newblock A feature subtraction method for image based kinship verification
  under uncontrolled environments.
\newblock In \emph{Image Processing (ICIP), 2015 IEEE International Conference
  on}, pages 1573--1577. IEEE, 2015.

\bibitem[Fang et~al.(2010)Fang, Tang, Snavely, and Chen]{fang2010towards}
Ruogu Fang, Kevin~D Tang, Noah Snavely, and Tsuhan Chen.
\newblock Towards computational models of kinship verification.
\newblock In \emph{Image Processing (ICIP), 2010 17th IEEE International
  Conference on}, pages 1577--1580. IEEE, 2010.

\bibitem[Fang et~al.(2013)Fang, Gallagher, Chen, and Loui]{fang2013kinship}
Ruogu Fang, Andrew~C Gallagher, Tsuhan Chen, and Alexander Loui.
\newblock Kinship classification by modeling facial feature heredity.
\newblock In \emph{Image Processing (ICIP), 2013 20th IEEE International
  Conference on}, pages 2983--2987. IEEE, 2013.

\bibitem[King(2009)]{dlib09}
Davis~E. King.
\newblock Dlib-ml: A machine learning toolkit.
\newblock \emph{Journal of Machine Learning Research}, 10:\penalty0 1755--1758,
  2009.

\bibitem[Kohli et~al.(2017)Kohli, Vatsa, Singh, Noore, and
  Majumdar]{kohli2017hierarchical}
Naman Kohli, Mayank Vatsa, Richa Singh, Afzel Noore, and Angshul Majumdar.
\newblock Hierarchical representation learning for kinship verification.
\newblock \emph{IEEE Transactions on Image Processing}, 26\penalty0
  (1):\penalty0 289--302, 2017.

\bibitem[Kohli et~al.(2018)Kohli, Yadav, Vatsa, Singh, and
  Noore]{kohli2018supervised}
Naman Kohli, Daksha Yadav, Mayank Vatsa, Richa Singh, and Afzel Noore.
\newblock Supervised mixed norm autoencoder for kinship verification in
  unconstrained videos.
\newblock \emph{IEEE Transactions on Image Processing}, 2018.

\bibitem[Kou et~al.(2015)Kou, Zhou, Xu, and Shang]{kou2015learning}
Lu~Kou, Xiuzhuang Zhou, Min Xu, and Yuanyuan Shang.
\newblock Learning a genetic measure for kinship verification using facial
  images.
\newblock \emph{Mathematical Problems in Engineering}, 2015, 2015.

\bibitem[Laiadi et~al.(2017)Laiadi, Ouamane, Benakcha, and
  Taleb-Ahmed]{laiadi2017rfiw}
Oualid Laiadi, Abdelmalik Ouamane, Abdelhamid Benakcha, and Abdelmalik
  Taleb-Ahmed.
\newblock Rfiw 2017: Lpq-sieda for large scale kinship verification.
\newblock In \emph{Proceedings of the 2017 Workshop on Recognizing Families In
  the Wild}, pages 37--39. ACM, 2017.

\bibitem[Li et~al.(2017)Li, Zeng, Zhang, Dai, Kan, Shan, and
  Chen]{li2017kinnet}
Yong Li, Jiabei Zeng, Jie Zhang, Anbo Dai, Meina Kan, Shiguang Shan, and Xilin
  Chen.
\newblock Kinnet: Fine-to-coarse deep metric learning for kinship verification.
\newblock In \emph{Proceedings of the 2017 Workshop on Recognizing Families In
  the Wild}, pages 13--20. ACM, 2017.

\bibitem[Liu et~al.(2016)Liu, Puthenputhussery, and Liu]{liu2016novel}
Qingfeng Liu, Ajit Puthenputhussery, and Chengjun Liu.
\newblock A novel inheritable color space with application to kinship
  verification.
\newblock In \emph{Applications of Computer Vision (WACV), 2016 IEEE Winter
  Conference on}, pages 1--9. IEEE, 2016.

\bibitem[L{\'o}pez et~al.(2016)L{\'o}pez, Boutellaa, and
  Hadid]{lopez2016comments}
Miguel~Bordallo L{\'o}pez, Elhocine Boutellaa, and Abdenour Hadid.
\newblock Comments on the “kinship face in the wild” data sets.
\newblock \emph{IEEE transactions on pattern analysis and machine
  intelligence}, 38\penalty0 (11):\penalty0 2342--2344, 2016.

\bibitem[Lopez et~al.(2018)Lopez, Hadid, Boutellaa, Goncalves, Kostakos, and
  Hosio]{lopez2018kinship}
Miguel~Bordallo Lopez, Abdenour Hadid, Elhocine Boutellaa, Jorge Goncalves,
  Vassilis Kostakos, and Simo Hosio.
\newblock Kinship verification from facial images and videos: human versus
  machine.
\newblock \emph{Machine Vision and Applications}, 29\penalty0 (5):\penalty0
  873--890, 2018.

\bibitem[Lu et~al.(2014)Lu, Zhou, Tan, Shang, and Zhou]{lu2014neighborhood}
Jiwen Lu, Xiuzhuang Zhou, Yap-Pen Tan, Yuanyuan Shang, and Jie Zhou.
\newblock Neighborhood repulsed metric learning for kinship verification.
\newblock \emph{IEEE transactions on pattern analysis and machine
  intelligence}, 36\penalty0 (2):\penalty0 331--345, 2014.

\bibitem[Lu et~al.(2017)Lu, Hu, and Tan]{lu2017discriminative}
Jiwen Lu, Junlin Hu, and Yap-Peng Tan.
\newblock Discriminative deep metric learning for face and kinship
  verification.
\newblock \emph{IEEE Transactions on Image Processing}, 26\penalty0
  (9):\penalty0 4269--4282, 2017.

\bibitem[Mahpod and Keller(2017)]{mahpod2017kinship}
Shahar Mahpod and Yosi Keller.
\newblock Kinship verification using multiview hybrid distance learning.
\newblock \emph{Computer Vision and Image Understanding}, 2017.

\bibitem[Moujahid and Dornaika(2018)]{moujahid2018pyramid}
A~Moujahid and F~Dornaika.
\newblock A pyramid multi-level face descriptor: application to kinship
  verification.
\newblock \emph{Multimedia Tools and Applications}, pages 1--20, 2018.

\bibitem[Nguyen and Bai(2010)]{nguyen2010cosine}
Hieu~V Nguyen and Li~Bai.
\newblock Cosine similarity metric learning for face verification.
\newblock In \emph{Asian conference on computer vision}, pages 709--720.
  Springer, 2010.

\bibitem[Patel et~al.(2017)Patel, Maheshwari, and Raman]{patel2017evaluation}
Bhavik Patel, RP~Maheshwari, and Balasubramanian Raman.
\newblock Evaluation of periocular features for kinship verification in the
  wild.
\newblock \emph{Computer Vision and Image Understanding}, 160:\penalty0 24--35,
  2017.

\bibitem[Qin et~al.(2015)Qin, Tan, and Chen]{qin2015tri}
Xiaoqian Qin, Xiaoyang Tan, and Songcan Chen.
\newblock Tri-subject kinship verification: Understanding the core of a family.
\newblock \emph{IEEE Transactions on Multimedia}, 17\penalty0 (10):\penalty0
  1855--1867, 2015.

\bibitem[Robinson et~al.(2016)Robinson, Shao, Wu, and Fu]{robinson2016families}
Joseph~P Robinson, Ming Shao, Yue Wu, and Yun Fu.
\newblock Families in the wild (fiw): Large-scale kinship image database and
  benchmarks.
\newblock In \emph{Proceedings of the 2016 ACM on Multimedia Conference}, pages
  242--246. ACM, 2016.

\bibitem[Robinson et~al.(2018)Robinson, Shao, Wu, Liu, Gillis, and
  Fu]{robinson2018visual}
Joseph~Peter Robinson, Ming Shao, Yue Wu, Hongfu Liu, Timothy Gillis, and Yun
  Fu.
\newblock Visual kinship recognition of families in the wild.
\newblock \emph{IEEE Transactions on Pattern Analysis and Machine
  Intelligence}, 2018.

\bibitem[Wang et~al.(2017)Wang, Robinson, and Fu]{wang2017kinship}
Shuyang Wang, Joseph~P Robinson, and Yun Fu.
\newblock Kinship verification on families in the wild with marginalized
  denoising metric learning.
\newblock In \emph{Automatic Face \& Gesture Recognition (FG 2017), 2017 12th
  IEEE International Conference on}, pages 216--221. IEEE, 2017.

\bibitem[Wang et~al.(2018)Wang, Ding, and Fu]{wang2018cross}
Shuyang Wang, Zhengming Ding, and Yun Fu.
\newblock Cross-generation kinship verification with sparse discriminative
  metric.
\newblock \emph{IEEE transactions on pattern analysis and machine
  intelligence}, 2018.

\bibitem[Weinberger and Saul(2009)]{weinberger2009distance}
Kilian~Q Weinberger and Lawrence~K Saul.
\newblock Distance metric learning for large margin nearest neighbor
  classification.
\newblock \emph{Journal of Machine Learning Research}, 10\penalty0
  (Feb):\penalty0 207--244, 2009.

\bibitem[Wu et~al.(2016)Wu, Boutellaa, L{\'o}pez, Feng, and
  Hadid]{wu2016usefulness}
Xiaoting Wu, Elhocine Boutellaa, Miguel~Bordallo L{\'o}pez, Xiaoyi Feng, and
  Abdenour Hadid.
\newblock On the usefulness of color for kinship verification from face images.
\newblock In \emph{Information Forensics and Security (WIFS), 2016 IEEE
  International Workshop on}, pages 1--6. IEEE, 2016.

\bibitem[Xia et~al.(2018)Xia, Xia, Zhou, Zhang, and Shao]{xia2018graph}
Chao Xia, Siyu Xia, Yuan Zhou, Le~Zhang, and Ming Shao.
\newblock Graph based family relationship recognition from a single image.
\newblock In \emph{Pacific Rim International Conference on Artificial
  Intelligence}, pages 310--320. Springer, 2018.

\bibitem[Xia et~al.(2011)Xia, Shao, and Fu]{xia2011kinship}
Siyu Xia, Ming Shao, and Yun Fu.
\newblock Kinship verification through transfer learning.
\newblock In \emph{Proceedings of the Twenty-Second International Joint
  Conference on Artificial Intelligence - Volume Volume Three}, IJCAI'11, pages
  2539--2544. AAAI Press, 2011.

\bibitem[Xu and Shang(2016)]{xu2016kinship}
Min Xu and Yuanyuan Shang.
\newblock Kinship verification using facial images by robust similarity
  learning.
\newblock \emph{Mathematical Problems in Engineering}, 2016, 2016.

\bibitem[Yan(2018)]{yan2018learning}
Haibin Yan.
\newblock Learning discriminative compact binary face descriptor for kinship
  verification.
\newblock \emph{Pattern Recognition Letters}, 2018.

\bibitem[Yan et~al.(2015)Yan, Lu, and Zhou]{yan2015prototype}
Haibin Yan, Jiwen Lu, and Xiuzhuang Zhou.
\newblock Prototype-based discriminative feature learning for kinship
  verification.
\newblock \emph{IEEE Transactions on cybernetics}, 45\penalty0 (11):\penalty0
  2535--2545, 2015.

\bibitem[Yang and Wu(2017)]{yang2017novel}
Yong Yang and Qingshan Wu.
\newblock A novel kinship verification method based on deep transfer learning
  and feature nonlinear mapping.
\newblock \emph{DEStech Transactions on Computer Science and Engineering},
  aiea, 2017.

\bibitem[Zhang et~al.(2016)Zhang, Xia, Pan, and Qin]{zhang2016genetics}
Junkang Zhang, Siyu Xia, Hong Pan, and AK~Qin.
\newblock A genetics-motivated unsupervised model for tri-subject kinship
  verification.
\newblock In \emph{Image Processing (ICIP), 2016 IEEE International Conference
  on}, pages 2916--2920. IEEE, 2016.

\bibitem[Zhang et~al.(2015{\natexlab{a}})Zhang, Huang, Song, Wu, and
  Wang]{zhang122015kinship}
Kaihao Zhang, Yongzhen Huang, Chunfeng Song, Hong Wu, and Liang Wang.
\newblock Kinship verification with deep convolutional neural networks.
\newblock In \emph{Proceedings of the British Machine Vision Conference
  (BMVC)}, pages 148.1--148.12. BMVA Press, September 2015{\natexlab{a}}.

\bibitem[Zhang et~al.(2015{\natexlab{b}})Zhang, Chen, and
  Saligrama]{zhang2015group}
Ziming Zhang, Yuting Chen, and Venkatesh Saligrama.
\newblock Group membership prediction.
\newblock In \emph{Proceedings of the IEEE International Conference on Computer
  Vision}, pages 3916--3924, 2015{\natexlab{b}}.

\bibitem[Zhao et~al.(2018)Zhao, Song, Zheng, and Shao]{zhao2018learning}
Yan-Guo Zhao, Zhan Song, Feng Zheng, and Ling Shao.
\newblock Learning a multiple kernel similarity metric for kinship
  verification.
\newblock \emph{Information Sciences}, 430:\penalty0 247--260, 2018.

\bibitem[Zhou et~al.(2016)Zhou, Shang, Yan, and Guo]{zhou2016ensemble}
Xiuzhuang Zhou, Yuanyuan Shang, Haibin Yan, and Guodong Guo.
\newblock Ensemble similarity learning for kinship verification from facial
  images in the wild.
\newblock \emph{Information Fusion}, 32:\penalty0 40--48, 2016.

\bibitem[Zhou et~al.(2018)Zhou, Jin, Xu, and Guo]{zhou2018learning}
Xiuzhuang Zhou, Kai Jin, Min Xu, and Guodong Guo.
\newblock Learning deep compact similarity metric for kinship verification from
  face images.
\newblock \emph{Information Fusion}, 2018.

\end{thebibliography}
\end{document}